\documentclass[conference]{IEEEtran}
\IEEEoverridecommandlockouts
\usepackage{cite}
\usepackage{amsmath,amssymb,amsfonts}
\usepackage{algorithmic}
\usepackage{booktabs}
\usepackage{graphicx}
\usepackage{stfloats}
\usepackage{textcomp}
\usepackage{multicol}
\usepackage{multirow}
\usepackage{xcolor}
\def\BibTeX{{\rm B\kern-.05em{\sc i\kern-.025em b}\kern-.08em
    T\kern-.1667em\lower.7ex\hbox{E}\kern-.125emX}}
\begin{document}

\title{Thinker: A vision-language foundation model for embodied intelligence
}

\author{\IEEEauthorblockN{
Baiyu Pan$^{*\dagger}$, Daqin Luo$^{*}$, Junpeng Yang, Jiyuan Wang, Yixuan Zhang, Hailin Shi$^{\dagger}$, 
Jichao Jiao$^{\dagger}$
}
\IEEEauthorblockA{
UBTECH Robotics, Shenzhen, China \\
Thinker@ubtrobot.com}
}

% \and
% \and
% \and
% \and
% \and

\maketitle
\renewcommand{\thefootnote}{}
\footnotetext{* Equal Contribution.}
\footnotetext{$\dagger$ Project Lead.}
\begin{abstract}
% When large vision language models(VLMs) are applied in the field of robotics, they encounter problems that are simple for humans yet error-prone for the VLMs. Such issues include confusion between the third-person and first-person perspective, and tends to focus on events in the middle while processing videos,  overlooking the information contained in the ending of the video.
When large vision-language models (VLMs) are applied in the field of robotics, they encounter problems that are simple for humans yet error-prone for the models. Such issues include confusion between third-person and first-person perspectives, and a tendency to overlook information in video endings during video reasoning. 
To address these challenges, we propose \textit{Thinker}, a large vision language foundation model designed for embodied intelligence. We tackle the aforementioned issues from two perspectives. 
% Firstly, we construct a large-scale dataset tailored for robot task planning, which encompasses ego-view videos, visual grounding and spatial understanding data, chain-of-thought data. 
Firstly, we construct a large-scale dataset tailored for robotic perception and reasoning, which encompasses ego-view videos, visual grounding, spatial understanding and chain-of-thought data. Secondly, we introduce a simple yet effective approach that substantially enhances the model’s capacity for video comprehension by jointly incorporating the key frame and video as inputs.
% video processing method, which integrates the key frame and video as input, thereby significantly enhances the models' ability to comprehend videos.
Our models achieved state-of-the-art results on two of the most commonly used benchmark datasets in the field of task planning.
% Finally, we integrate reinforcement learning with the proposed graph-based task planning scheme, thereby providing a broader explore space for reinforcement learning.

\end{abstract}


\section{Introduction}

Recently, large vision language models (VLMs) have achieved remarkable results across a wide range of domains. This has led numerous researchers to adopt VLMs in the field of robotics\cite{ji2025robobrain,liu2024robomamba,team2025robobrain}. While VLMs excel at scene understanding, they face significant challenges in planning. In particular, they struggle to predict the robot's future state based on the current and past observations.
Most VLMs are trained primarily on visual question answering (VQA) and image captioning datasets\cite{schuhmann2021laion,xu2024llava, kazemzadeh2014referitgame}, where scenes are typically depicted from a third-person perspective. The absence of robot-specific training data fundamentally constrains the capacity of current models to enable effective robotic task planning.

The fundamental reason for the gap between existing VLMs and their application in robotics is the absence of temporal and spatial information grounded in a first-person perspective. To address this limitation, we introduce our foundation model, \textit{Thinker}, along with its four core capabilities: 
\textbf{Task Planning} capability enables the VLMs to understand user instructions, maintaining memory of past states and predict future states. 
\textbf{Spatial Understanding} capability allow the VLMs to establish an egocentric coordinate system, with the camera as the origin, such that all states and spatial relationships are defined relative to this frame of reference.
\textbf{Temporal Understanding} capability is essential for VLMs to extract key information from past events and integrate this historical information with current instructions to assess the present state.
\textbf{Object Grounding} capability enables the VLMs to describe the object with a format of bounding box and points.
Accordingly, we have constructed a large-scale dataset to foster these capabilities. Comprehensive experiments demonstrate that \textit{Thinker} exhibits strong general and robotic capabilities.

In this paper, we reveal partial details of the \textit{Thinker}, including training data, model architecture, training methods, and evaluation results. We also report that our model has achieved state-of-the-art (SOTA) results on the Robovqa\cite{sermanet2024robovqa} and Egoplan-bench2\cite{qiu2024egoplan} benchmarks.

\begin{figure*}[htbp]
\centerline{
\includegraphics[width=0.3\linewidth]{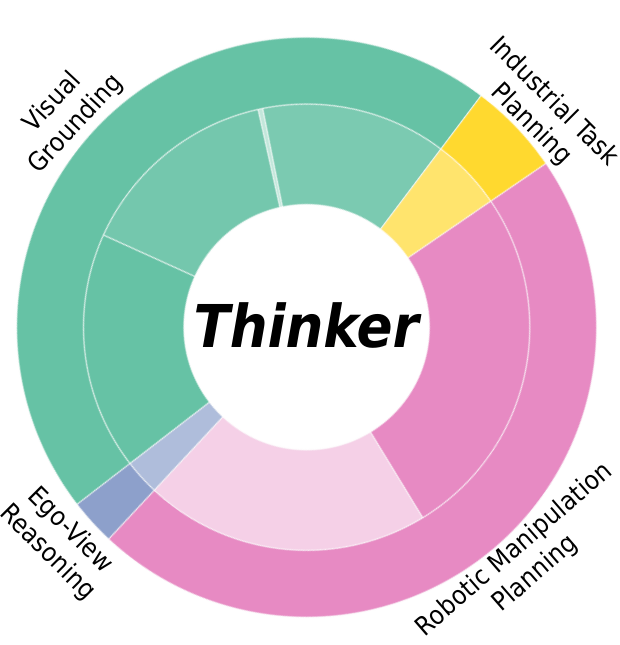}
\includegraphics[width=0.6\linewidth]{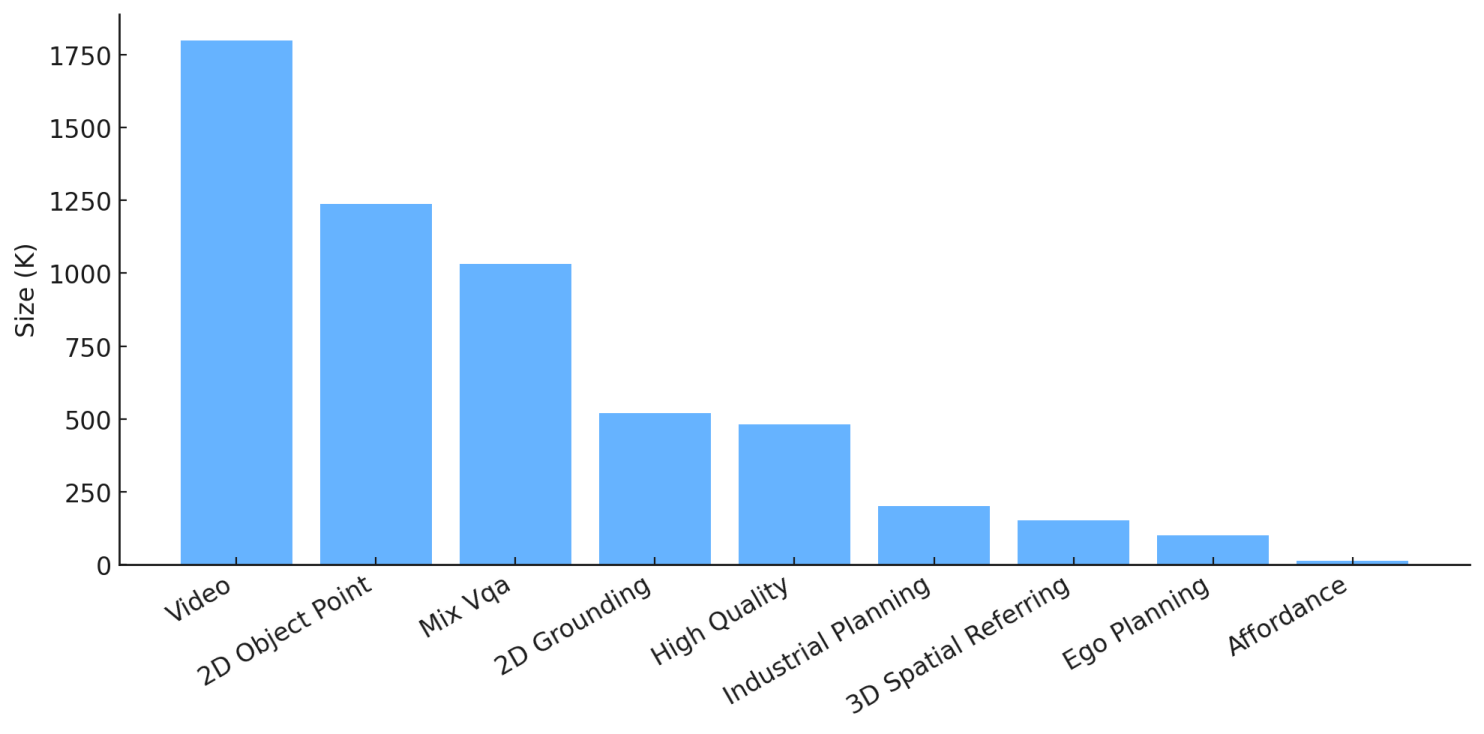}
}
\caption{
This figure illustrates the distribution of our crafted training datasets, that we categorized in to four types: visual grounding, ego-view, planning, industrial.
% including interactive reasoning with long-horizon planning and closed-loop feedback, spatial perception for precise point and bounding box prediction from complex instructions, and multi-agent collaboration tasks, which is meticulously categorized into three primary types: 
}
\label{fig_static}
\end{figure*}

% Our architecture encodes multi-resolution images, video frames, language instructions into a token sequence for unified processing.

\section{Training Data}
\label{sec:train data}

% To enhance spatiotemporal understanding and long-horizon planning capabilities, 

% We have crafted four kinds of datasets, as shown in Table~\ref{fourdatasets}: the spatial, temporal understanding, ego-view reasoning, planing, and our in-house industrial-specif dataset, \textit{Industroplan}, which focuses on multi-object manipulation and transport tasks in industrial environments.
As shown in Fig~\ref{fig_static}, Thinker is trained on large-scale and diverse datasets that strengthen its capabilities in embodied settings. Specifically, we crafted four kinds of datasets covering spatial and temporal understanding, ego-view reasoning, planning, and our in-house industrial dataset, \textit{Industroplan}, which focuses on multi-object manipulation and transport tasks in industrial environments. A summary is provided in Table~\ref{fourdatasets}.

\begin{table}[h]
\caption{Overview of the four constructed datasets.}
\centering
\begin{tabular}{l|l|l}
\toprule
Constructed datasets& Originates    &   Size* \\\midrule
\multirow{4}{*}{Visual Grounding}       & 
Lvis-520K,
& \multirow{4}{*}{1.7M}  \\
&Sharerobot-affordance-6.5K, & \\
& Pixmopoint-570K,&  \\
&Robopoint-667K &  \\\midrule
% & & &  \\\midrule
Ego-View Reasoning  & Egoplan-it-100K       & 100K  \\\midrule
\multirow{2}{*}{Robotic Manipulation Planning} & Robovqa-800K, & \multirow{2}{*}{1.8M}    \\
& Sharerobot-1M &  \\\midrule
Industrial Task Planning      & Industroplan-200K   & 200K   \\\bottomrule
\multicolumn{3}{l}{\small * the number of files}
\end{tabular}
\label{fourdatasets}
\end{table}

% Building on this, we further construct the IndustroPlan dataset, which focuses on multi-object manipulation and transport tasks in industrial environments. It substantially improves the model’s long-horizon reasoning and corrective capabilities in real-world scenarios.
% Our datasets include:
% In the following, we describe four key components of our training data.
% Spatial Understanding Data, Ego-View Planning Data, Robotic Manipulation Planning Data, and Industrial Task Planning Data.

\subsection{Visual Grounding Data}
\label{subsec:Spatial Understanding Data}
To develop robust spatial perception, we construct visual grounding datasets for both bounding box and point-level object localization. For bounding box grounding, We built Lvis-520K based on \cite{ramanathan2023paco} which includes QA pair generated by GPT-4o\cite{hurst2024gpt} regarding object's functions (e.g., \emph{“Which part of the bicycle is responsible for steering?”}). We also train the model to learn graspable regions by leveraging the Sharerobot-affordance-6.5K\cite{ji2025robobrain}. For point grounding, we utilized a refined version of Pixmopoint-570K\cite{deitke2025molmo} and Robopoint-667K\cite{yuan2024robopoint}, which we remove instances with more than 10 points and outdoor scenes. 
These datasets collectively supports the development of precise spatial perception and instruction understanding.

\subsection{Ego-View Reasoning Data}
\label{subsec:Ego-View}
We constructed Egoplan-it-100K by carefully filtering and refining Egoplan-it\cite{chen2023egoplan}, with the aim of advancing temporal reasoning and egocentric task planning.
Each item includes a video clip and the last frame. We designed two task formats: open-ended and multiple-choice question. We use the labeled action as the correct option and randomly sample at least three actions from other sequences as distractors for the multiple-choice question.

\subsection{Robotic Manipulation Planning Data}
\label{subsec:Robotic}
We construct a large-scale robotic planning dataset, Robovideo-1.8M, by integrating Robovqa~\cite{sermanet2024robovqa} and Sharerobot~\cite{ji2025robobrain}. Robovqa\cite{sermanet2024robovqa} is a large-scale dataset comprising over 800K QA pairs that span multiple embodiments, including robots, humans, and tool-assisted human interactions. In contrast, Sharerobot\cite{ji2025robobrain} contains 1M QA pairs designed for fine-grained planning in robotic manipulation tasks, covering 102 diverse scenes and 12 robot embodiments derived from Open-x-embodiment \cite{o2024open}.
By training with Robovideo-1.8M, \textit{Thinker} acquires the ability to spontaneously perform complex reasoning in robotic task scenarios.

\subsection{Industrial Task Planning Data}
\label{subsec:Industrial}
To further strengthen long-horizon reasoning in real-world scenarios, we construct the \textit{Industroplan}-200K dataset, which focuses on task planning in industrial environments involving multi-object manipulation and transportation. Each instance includes video demonstrations, task goals, and chain-of-thought annotations, covering diverse layouts, object types, and action sequences. \textit{Industroplan} is explicitly designed for long-horizon tasks, making it suitable for training and evaluating robotic perception and planning in complex factory environments.
% This dataset complements RoboVideo-1.8M and EgoPlan-IT-100K, extending coverage to domain-specific reasoning in structured industrial settings.

\begin{figure*}[htbp]
\centerline{\includegraphics[width=0.7\linewidth]{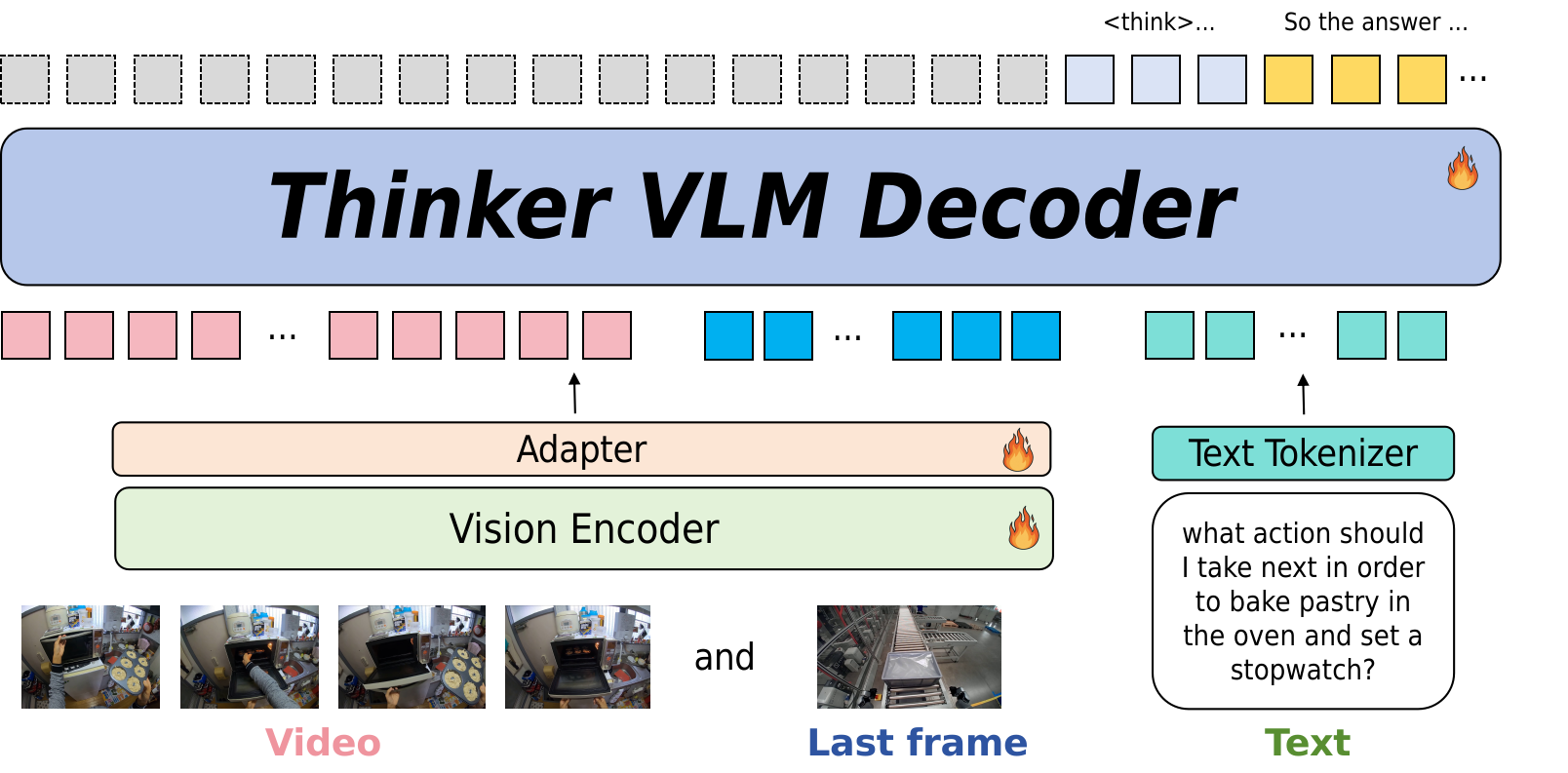}}
\caption{Our model supports images, videos, and complex language instructions. Video sequences, along with their last frames, are processed through a vision encoder and an adapter. All input tokens are subsequently concatenated and fed into the \textit{Thinker} decoder.
}
\label{fig_workflow}
\end{figure*}

\section{Thinker Models}
\subsection{Model architecture}
We have developed the \textit{Thinker} base model, a large vision language model with ten billion level parameters. The architecture is shown in Fig \ref{fig_workflow}. \textit{Thinker}  comprises four modules: a text tokenizer, a visual encoder and a multi-layer perceptron to align the visual and language space, and a language model backbone. This design achieves a unified representation across vision, language, and time. This allows \textit{Thinker} to accurately capture visual details, comprehend task instructions, and conduct reasoning in multiple scenarios, thereby providing a reliable foundation for embodied intelligence.

\subsection{Training Strategy}
Thinker is trained with a two-step strategy to develop robust task planning in complex scenarios. 
In the first stage, we focus on building the model’s foundational perception and reasoning capabilities. 
In the second stage, we perform supervised fine-tuning on downstream planning tasks to align its reasoning capability with task-specific goals. 
This strategy empowers the model to extend its reasoning to diverse scenarios, adapt to downstream tasks, and ultimately produce executable plans in real-world settings.

\subsection{Stage-1: Building Embodied Capabilities}
\label{sec:stage1}
The first stage focuses on establishing \textit{Thinker}’s foundational embodied capabilities. We fine-tune \textit{Thinker} on a combination of general datasets, spatial understanding datasets, and large-scale planning datasets, which equips it with robust spatial perception and reasoning skills, thereby providing a solid foundation for downstream task alignment and long-horizon planning in complex scenarios. In addition, we incorporate the last frame of each video clip as an auxiliary input during video understanding training, which further enhances the model’s performance.

\subsection{Stage-2: Downstream Task Fine-Tuning}
\label{sec:stage2}

The second stage focuses on aligning \textit{Thinker}'s reasoning capabilities with complex industrial planning tasks. We perform supervised fine-tuning on the Industroplan-200K dataset. This process enables the model to adapt its inherited reasoning ability form Stage-1 to sequential dependencies, diverse object layouts, and corrective feedback. As a result, \textit{Thinker} can generate executable plans in real-world industrial scenarios, effectively bridging the spatial understanding with practical task execution.

\begin{table*}[htbp]
\centering
\caption{Performance comparison of different models on RoboVQA and EgoPlan-Bench2 benchmarks. Best results are in bold, and the second-best results are underlined.}
\label{tab:robovqa_egoplan}
\begin{tabular}{l|lllll|lllll}
\toprule
& \multicolumn{5}{c}{Robovqa}                  & \multicolumn{5}{|c}{Egoplan-bench2}                 \\
                & BLEU-1 & BLEU-2 & BLEU-3 & BLEU-4 & BLEU-avg & Daily life & Work  & Recreation & Hobbies & Overall \\\midrule
Qwen2.5-VL-7B\cite{bai2025qwen2}     & 62.2          & 54.6          & 48.7          & 45.0          & 52.6          & 31.4           & 26.7           & 29.5           & 28.6           & 29.1           \\
GPT-4V\cite{achiam2023gpt}            & 32.2          & 26.5          & 24.7          & 23.9          & 26.8          & 36.7           & 27.7           & 33.9           & 32.5           & 32.6           \\
Cosmos-Reason1-7B\cite{azzolini2025cosmos} & /             & /             & /             & /             & /             & 30.7           & 31.6           & 20.3           & 20.3           & 26.8           \\
ThinkAct-7B\cite{huang2025thinkact}       & 69.1          & 61.8          & 56.0          & 52.4          & 59.8          & 50.1           & 49.8           & 44.8           & 45.2           & 48.2           \\
% Robix-7B-Base[7]     & /             & /             & /             & /             & 53.6          & /              & /              & /              & /              & /              \\
% Robix-32B-Base[7]    & /             & /             & /             & /             & 48.3          & /              & /              & /              & /              & /              \\
% VeBrain-8B        & /             & /             & /             & /             & /             & 31.79          & 35.31          & 31.19          & 34.43          & 27.30          \\
% Magma-8B          & /             & /             & /             & /             & /             & 4.56           & 3.39           & 6.56           & 2.97           & 4.09           \\
% Cosmos-Reason1-7B & /             & /             & /             & /             & /             & 30.75          & 27.12          & 31.69          & 20.30          & 26.87          \\
% ThinkAct-7B       & 69.1          & 61.8          & 56.0          & 52.4          & 59.8          & 50.1           & 49.8           & 44.8           & 45.2           & 48.2           \\
% Robomaba          & 54.9          & 44.2          & 39.5          & 36.3          & 43.72         & /              & /              & /              & /              & /              \\
RoboBrain-7B\cite{ji2025robobrain}         &\underline{72.05}   & \underline{65.35}   & \underline{59.39}   & \underline{55.05}   & \underline{62.7}    & /              & /              & /              & /              & /              \\
RoboBrain2-7B\cite{team2025robobrain}     & 37.4             & 31.0             & 27.1            & 25.8         & 30.0             & 39.41          & 32.20          & 33.88          & 26.98          & 33.23          \\
RoboBrain2-32B\cite{team2025robobrain}    & /             & /             & /             & /             & /             & \textbf{64.01} & \underline{53.22}    & \underline{57.92}    & \underline{52.48}    & \underline{57.23}    \\\midrule
\textbf{\textit{Thinker-7B}}      &  \textbf{72.7} & \textbf{65.7} & \textbf{59.5} & \textbf{56.0} & \textbf{63.5} & \underline{63.78}    & \textbf{54.95} & \textbf{61.20} & \textbf{52.54} & \textbf{58.21}\\\bottomrule
\end{tabular}
\end{table*}

\section{Infrastructures}
\label{sec:infrastructures}
This section outlines our infrastructure that supports the training, fine-tuning, and inference of \textit{Thinker}. The stack is built to (i) train jointly on heterogeneous datasets, (ii) perform parameter-efficient fine-tuning on \textit{\textbf{Thinker-7B}}, one of our proposed models, and (iii) deploy under benchmark protocols with reliability and observability.

\subsection{Large-Scale Multi-Task Training Infrastructure}
\label{subsec:multi_task_training_infra}

% \paragraph{Design Goals}
We address three practical challenges in multi-task, multi-modal training: (1) heterogeneity across sources (video with temporal context vs. single-image VQA), (2) efficient and reproducible initialization from a large pre-trained backbone, and (3) stable throughput at scale.
% \paragraph{Unified Multi-Source Pipeline}
We adopt a \emph{unified sampling schema} that normalizes all examples into a task-aware structure covering visual inputs, textual inputs, supervision targets, and task type. Balanced task mixing is implemented with a \emph{dynamic sampler} that adapts to validation feedback, ensuring both datasets contribute meaningfully during training. 
% To reduce I/O pressure and improve device utilization, lightweight metadata checks are performed offline, while \emph{deferred preprocessing} (visual transforms and tokenization) occurs just-in-time within the training step. 
Moreover, we employ \emph{sharded loading} and \emph{selective freezing} to minimize memory pressure and warm-up time. 
% The procedure yields reproducible startup behavior and leaves sufficient headroom for long-context batches on GPUs.

% \subsection{Fine-Tuning Infrastructure}
% \label{subsec:lora_finetuning_infra}

% \paragraph{Adapter Placement}
% We fine-tune the model with \emph{LoRA adapters} while keeping the pre-trained backbone frozen. Adapters are placed to align with language–vision fusion capacity in the generative pathway, enabling the model to specialize for embodied planning and VQA without modifying core perception modules.

% \paragraph{Stable and Efficient Training}
% To balance efficiency and stability, we combine \emph{mixed-precision training} with \emph{activation checkpointing} and strict gradient isolation for frozen parameters. This recipe keeps memory usage predictable for long sequences and maintains stable optimization behavior across multi-task updates, without exposing backbone internals or hyperparameter minutiae.

\subsection{Inference Infrastructure for Fine-Tuned Model}
\label{subsec:inference_infra}

% \paragraph{Merge-and-Serve Deployment}
% For deployment, LoRA adapters are \emph{merged offline} into the backbone to produce a single consolidated checkpoint. We then apply \emph{hardware-aware compilation and scheduling} to reduce latency and increase throughput in both datacenter and edge-style settings, while preserving numerical consistency with the fine-tuned model.

% \paragraph{Benchmark-Adaptive Pipeline}
A \emph{task-aware inference pipeline} Standardizes inputs and outputs for EgoPlan-Bench2 and RoboVQA. Video inputs are converted to concise temporal visual representations for planning, whereas static-image VQA inputs are formatted for compact reasoning. Outputs are normalized to comply with each benchmark’s evaluation protocol, enabling seamless and repeatable assessment.

\subsection{Fault Tolerance and Monitoring}
\label{subsec:fault_tolerance}

% \paragraph{Resilience}

% \paragraph{Observability}
We continuously track optimization signals (per-task losses), throughput, accelerator memory, and device utilization. Automated alerts surface anomalies (e.g., drops in utilization or loss drift), enabling rapid operator intervention with minimal wasted compute.
Long-horizon training runs employ \emph{periodic checkpointing} (model, optimizer, and data-loader cursor) to allow swift recovery from node failures. On interruption, the launcher resumes from the latest consistent state without reprocessing large portions of the dataset.

% \subsection{Summary}
% \label{subsec:infra_summary}
% Our infrastructure delivers a concise, reproducible path from multi-source data to deployment for embodied vision–language modeling: unified multi-task training across EgoPlan-Bench2 and RoboVQA, LoRA-based adaptation on \textbf{Qwen2.5-VL-Instruct-7B} with stable large-scale optimization on \textbf{32 H200 GPUs}, and a merge-and-serve inference stack tailored to benchmark protocols. The result is an efficient, reliable system that supports end-to-end experimentation and deployment without exposing low-level implementation details.

\section{Evaluation  Results}

% \begin{figure*}[!t]
%     \centering
%     \includegraphics[width=\textwidth]{images/robovqa_egoplan.pdf} % 跨双栏
%     \caption{The performance of our model \textit{Thinker-7B} on the RoboVQA and EgoPlan-Bench2 benchmarks. \textit{Thinker-7B} surpassed all baseline models, achieving state-of-the-art results.}
%     \label{fig:robovqa_egoplan}
% \end{figure*}

\subsection{Benchmarks and Protocols}
\label{sec:benchmarks and protocols}
% To comprehensively validate the effectiveness of our proposed model, 
For evluation on \textbf{Robovqa}\cite{sermanet2024robovqa},
we report BLEU-1$\sim$4 on free-form textual answers as our primary metric and treat a match against any reference in the answer set as correct.
We adopt the standard Top-1 accuracy as the primary evaluation metric for \textbf{Egoplan-bench2}\cite{qiu2024egoplan}. 
These datasets collectively cover a broad spectrum of video-language reasoning and planning capabilities, ensuring a rigorous and fair assessment of performance.

% These domains are further broken down into 24 fine-grained scenarios, allowing for nuanced evaluation across diverse activity contexts. Each instance is structured to provide: (i) a natural language goal description, (ii) a short \emph{task-progress} clip representing historical context, and (iii) a current observation frame. The model must then infer the appropriate \emph{next action} from four candidate choices.
% Given its multiple-choice nature, 
% This setup directly reflects a model’s ability to make precise action predictions in real-world planning scenarios. 
% Compared with RoboVQA, EgoPlan-Bench2 emphasizes decision-making under uncertainty and temporal reasoning grounded in egocentric perspectives, making it a valuable complement to the RoboVQA benchmark.

\subsection{Main Results}
\label{sec:main results}

We compare our proposed \textit{Thinker-7B} against seven SOTA VLM baselines that span both open-source and closed-source families: Qwen2.5-VL-7B~\cite{bai2025qwen2}, GPT-4V~\cite{achiam2023gpt}, Cosmos-Reason1-7B~\cite{azzolini2025cosmos}, ThinkAct-7B~\cite{huang2025thinkact}, RoboBrain-7B~\cite{ji2025robobrain}, RoboBrain2-7B~\cite{team2025robobrain}, and RoboBrain2-32B~\cite{team2025robobrain}. 
Table \ref{tab:robovqa_egoplan} reports the full results on \textit{Robovqa} and \textit{Egoplan-bench2}.

% Our \textit{Thinker-7B} achieves the best scores on all BLEU variants, with BLEU-1/2/3/4 of \textbf{72.7, 65.7, 59.5, 56.0}. 
% Averaged over BLEU-1$\sim$4, \textit{Thinker-7B} attains \textbf{63.5}, exceeding prior strong systems including RoboBrain-7B (62.7; +0.8), ThinkAct-7B (59.8; +3.7), Qwen2.5-VL-7B (52.6; +10.9), and GPT-4V (26.8; +36.7). 
% This result highlights its capability to decompose complex long-range task planning.

\paragraph{Performance on Robovqa}  
\textit{Thinker-7B} achieves the best performance across BLEU-1/2/3/4 with scores of \textbf{72.7, 65.7, 59.5, 56.0} and a average BLEU score of \textbf{63.5}. Specifically, \textit{Thinker-7B} surpass the second best\cite{ji2025robobrain} by \textbf{0.8}. 
This results highlight \textit{Thinker-7B}’s ability to parse fine-grained spatiotemporal cues and to decompose complex long-range planning tasks into coherent textual responses. Importantly, the margin over GPT-4V demonstrates the necessity of robotics-tailored training, as general-purpose VLMs struggle to align with task-specific reasoning requirements.

% On the egocentric next-action planning benchmark, \textit{Thinker-7B} delivers a new state of the art with an overall Top-1 accuracy of \textbf{58.21}, this surpasses all general and embodied baselines we evaluate, including ThinkAct-7B (48.2; +10.01), RoboBrain2-7B (33.23; +24.98), GPT-4V (32.6; +25.61), Qwen2.5-VL-7B (29.1; +29.11), and Cosmos-Reason1-7B (26.8; +31.41). With domain-wise breakdown Daily life: \textbf{63.78}, Work: \textbf{54.95}, Recreation: \textbf{61.20}, and Hobbies: \textbf{52.54}. Notably, despite having only 7B parameters, \textit{Thinker-7B} also outperforms the larger RoboBrain2-32B.

\paragraph{Performance on Egoplan-bench2}  
With the best accuracy of 58.2, \textit{Thinker-7B} outperforms all baselines in both general and embodied VLMs comprehensively.
The consistently strong scores across diverse domains, first place in three of four, indicate that our model is not only adept at handling common household or recreational tasks but also exhibits competitive planning ability in professional and work-related scenarios. This breadth of capability confirms \textit{Thinker-7B}’s adaptability to diverse egocentric contexts.

Across two challenging robotic benchmarks, our model consistently outperforms both general-purpose and embodied VLMs, highlighting its robust video perception and robot-task understanding capabilities.

\section{Future works}
We will soon release the full technical report of \textit{Thinker}, along with detailed descriptions, and open-source its architecture and weights. In parallel, we plan to explore world models and video–language–action models built upon this foundation.

% \section*{References}

% Please number citations consecutively within brackets \cite{greenwade93}. The 
% sentence punctuation follows the bracket \cite{greenwade93}. Refer simply to the reference 
% number, as in \cite{greenwade93}---do not use ``Ref. \cite{greenwade93}'' or ``reference \cite{greenwade93}'' except at 

\bibliographystyle{IEEEtran}
% \bibliography{ref}
\bibliography{ref_new}

% \color{red}
% IEEE conference templates contain guidance text for composing and formatting conference papers. Please ensure that all template text is removed from your conference paper prior to submission to the conference. Failure to remove the template text from your paper may result in your paper not being published.

\end{document}